\documentclass[letterpaper, 10 pt, conference]{ieeeconf}  

\IEEEoverridecommandlockouts                              

\usepackage{graphics} 
\usepackage{epsfig} 
\usepackage{mathptmx} 
\usepackage{times} 
\usepackage{amsmath} 
\usepackage{amssymb}  
\usepackage[inkscapelatex=false]{svg}
\usepackage{graphicx}
\usepackage{makecell}
\usepackage{multirow}
\usepackage{color}
\usepackage[hidelinks]{hyperref}
\usepackage[hyphenbreaks]{breakurl}
\usepackage{cite}

\title{\LARGE \bf
{\color{cyan}{STOP}}Net: Multiview-based 6-DoF {\color{cyan}{S}}uction Detection for\\
{\color{cyan}{T}}ransparent {\color{cyan}{O}}bjects on {\color{cyan}{P}}roduction Lines
}

\author{Yuxuan Kuang$^{1,2*}$, Qin Han$^{1*}$, Danshi Li$^{2,3}$, Qiyu Dai$^{1}$, Lian Ding$^{4}$, Dong Sun$^{4}$, Hanlin Zhao$^{4}$, He Wang$^{1 \dag}$ 
\thanks{$^{1}$Center on Frontiers of Computing Studies, School of Computer Science, Peking University.}%
\thanks{$^{2}$Galbot.}%
\thanks{$^{3}$New York University.}%
\thanks{$^{4}$Huawei Cloud Computing Technologies Co., Ltd.}%
\thanks{*The first two authors contributed equally.}%
\thanks{\dag Corresponding to hewang@pku.edu.cn.}%
}

\begin{document}

\maketitle
\thispagestyle{empty}
\pagestyle{empty}

\begin{abstract}

In this work, we present STOPNet, a framework for 6-DoF object suction detection on production lines, with a focus on but not limited to transparent objects, which is an important and challenging problem in robotic systems and modern industry. Current methods requiring depth input fail on transparent objects due to depth cameras' deficiency in sensing their geometry, while we proposed a novel framework to reconstruct the scene on the production line depending only on RGB input, based on multiview stereo. Compared to existing works, our method not only reconstructs the whole 3D scene in order to obtain high-quality 6-DoF suction poses in real time but also generalizes to novel environments, novel arrangements and novel objects, including challenging transparent objects, both in simulation and the real world. Extensive experiments in simulation and the real world show that our method significantly surpasses the baselines and has better generalizability, which caters to practical industrial needs.

\end{abstract}
\section{INTRODUCTION}

Object picking on production lines is an essential task for robotic systems and is widely used in modern industrial applications, such as logistics sorting and bin picking, \textit{etc.} Building such an autonomous robotic system to pick and place objects on a moving production line can greatly save time, reduce human labor and increase productivity. In addition, as a major end-effector in picking objects, the suction cup has gained much popularity due to its simplicity, efficiency and robustness. Therefore reliable suction detection is also worth studying to solve this task.

Due to the importance of this task, recent years have witnessed great progress in developing such systems to automatically conduct production line pick and place~\cite{cad_matching,industry4.0,assembly,pick_by_vision,morph,practical,akinola2021dynamic}. These methods rely on low-level image features or 3D CAD models to detect suction poses, which only work on regular or seen objects. There are also learning-based methods on static object suction~\cite{andy-affordance,mahler2018dexnet,dexnet4.0,suctionnet,RGBD-seg-sucdetect} that leverage accurate surface information from RGBD/depth input and can generalize to novel diffuse objects.

However, transparent object suction still remains a challenging case for such a system. Depth sensors can hardly sense transparent objects and often generate wrong, missing and fuzzy depth on these objects, which causes the failure of these learning-based methods on transparent objects.

\begin{figure}
\centering
\includegraphics[width=1.0\linewidth]{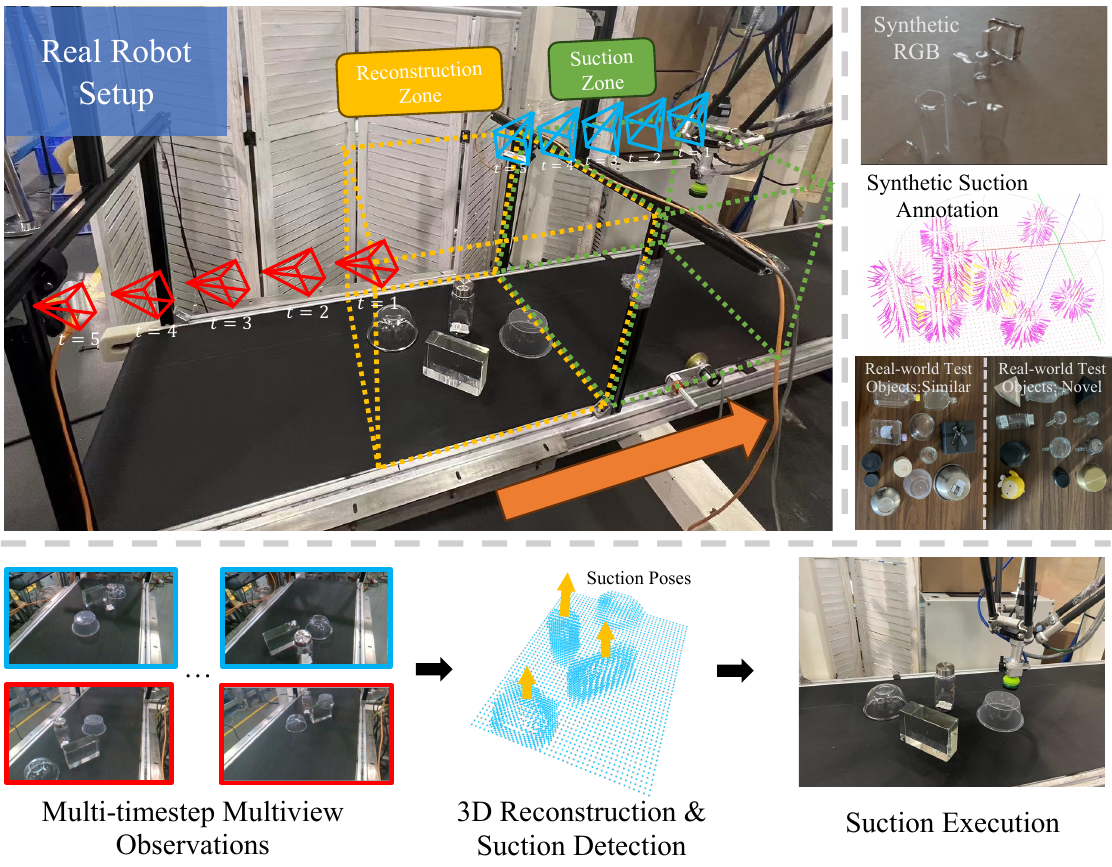}
\caption{Overview of our production line real robot setup and our proposed STOPNet. Taking multi-timestep RGB input from two cameras, our framework conducts 3D scene reconstruction and suction detection in real time. Our model is trained on a large-scale synthetic dataset but it can also generalize to real-world objects and environments effectively.
}
\vspace{-15pt}
\label{fig/teaser}
\end{figure}

For grasping transparent objects, recent studies have focused on depth restoration~\cite{sajjan2019cleargrasp,fang2022transcg,dai2022domain}, leveraging imperfect depth~\cite{Weng-2020-123091,soft} and RGB-based grasping~\cite{zhou2019glassloc,ghostpose,Dai2023GraspNeRF}. However, recent works for suction detection~\cite{andy-affordance,mahler2018dexnet,dexnet4.0,suctionnet,RGBD-seg-sucdetect} require accurate depth input, since suction relies more on local surface geometry than 6-DoF grasping. Therefore these methods struggle to detect suction on transparent objects without reliable depth input. In that sense, the issue of suction detection for transparent objects is worthy of study.

As to production line pick and place system design, recent works mainly focus on suction detection from a single view~\cite{cad_matching,industry4.0,morph}. These methods usually suffer from performance degradation in cluttered scenes due to their limited observations. Instead, our method utilizes multi-timestep multiview images to reconstruct the whole scene using volumetric truncated signed distance function (TSDF) representation, making it easier to handle cluttered scenes, which are common in real-world applications.

Our method, STOPNet, addresses the problems mentioned above by ``stopping" the production line. Specifically, assuming that the production line is moving at a constant speed, we can transform our multi-timestep multiview RGB observations to static multiview observations, which enable 3D reconstruction. After that, we tackle this problem by combining 2D and 3D, where in 3D space, we conduct scene reconstruction and wrench-collision prediction only given multiview RGB images, and in 2D space, we predict pixel-level seal and normal on original RGB input since these two attributes rely more on local object surface smoothness than global scene geometry. By conducting the 2D-3D fusion, we can obtain reliable Top-\textit{k} suction poses, which will be further executed by robot suction cups.

To achieve generalizability and reduce the sim2real gap while saving training costs, we utilize a domain randomization-based synthetic data generation pipeline to generate a large-scale and diverse synthetic dataset containing over 300K images and 40M suction poses for our training. Combining this with other designs, such as discarding redundant textures and illuminations, extensive experiments demonstrate our method's generalizability both in simulation and the real world.

To summarize, our framework has many advantages over current methods on production line object suction: it is generalizable to diverse objects in the real world, achieving success rates up to 90.38\% in the real world; it performs well on transparent objects without depth input, which is a challenging case in practice; it utilizes multi-timestep multiview input to conduct 3D reconstruction, thus is able to handle cluttered scenes; it is in real time and can be adapted to practical industrial production; it has a low cost of only 2 RGB cameras; and it can perform 6-DoF suction detection.
\section{RELATED WORK}

\subsection{Transparent Object Grasping}

Grasping transparent objects is a challenging task for robotic manipulation since stereo cameras can hardly capture good-quality depth images of transparent objects. Thus recent studies on object suction~\cite{andy-affordance,mahler2018dexnet,dexnet4.0,suctionnet,RGBD-seg-sucdetect} based on RGBD or depth images will inevitably fail on transparent objects due to their imperfect depth.

Recent years have witnessed multiple solutions to this issue. One direct way is to restore depths in advance, such as in~\cite{sajjan2019cleargrasp,fang2022transcg,dai2022domain}. Some studies focus on using single-view fuzzy RGBD images to learn a grasping policy~\cite{Weng-2020-123091,soft}, which cannot handle cluttered scenes due to their single-view input. There are also works focusing on taking multiview RGB as input. For example, GlassLoc~\cite{zhou2019glassloc} constructs a Depth Likelihood Volume (DLV) descriptor from multiview light field observations to represent transparent object clutter scenes. GhostPose~\cite{ghostpose} conducts transparent object grasping by a model-free pose estimation method based on multiview geometry. And GraspNeRF~\cite{Dai2023GraspNeRF} represents the scene geometry as the generalizable NeRF and jointly trains the NeRF and grasping detection. Similar to~\cite{Dai2023GraspNeRF}, our method predicts TSDF and detects suction poses directly from multiview RGB images, thus being able to handle transparent objects.

\subsection{Suction Analytic Models and Suction Detection}

Different from grasping by parallel grippers, suction cups have the advantage of high efficiency, flexibility and simplicity, thus are widely used in industrial production lines. Many studies~\cite{mahler2018dexnet,suction_cup_modeling,suctionnet} have proposed suction analytic models and their evaluation. As elaborated in~\cite{suctionnet}, a suction pose is defined as a 3D suction point and a direction vector starting from the suction point and pointing outside of the object surface. And a suction pose can be evaluated in dimensions of seal, wrench and collision. We follow their suction analytic model in our setting.

As to suction detection, database-based or CAD model-based suction detection~\cite{cad_matching} is one of the most commonly used methods in the industry since it's simple and efficient. However, it is not generalizable and thus cannot handle a wide variety of objects. Current learning-based methods, such as~\cite{andy-affordance,mahler2018dexnet,dexnet4.0,suctionnet,RGBD-seg-sucdetect} are generalizable but only cater to single view static suction and need depth input, which will fail on transparent objects. Our method, instead, not only utilizes multi-timestep multiview RGB images to handle dynamic and cluttered scenes but is also generalizable to a wide variety of objects, including challenging transparent objects.

\subsection{Object Picking on Production Lines}

In line with the principles of Industry 4.0~\cite{Lasi2014} which aims to push for flexibility on target changes and autonomy, object picking on production lines has wide applications in autonomous industrial scenarios, thus was heavily studied in the previous decades~\cite{cad_matching,industry4.0,assembly,pick_by_vision,morph,practical,akinola2021dynamic}. Recent studies implement object recognition and suction detection mainly by CAD model database matching~\cite{cad_matching}, HOG/SVM~\cite{industry4.0} or morphological operators~\cite{morph}, which are either non-generalizable or ineffective. By contrast, our method can both generalize to novel objects and achieve high performance, with a low cost of 2 RGB cameras.
\section{METHOD} \label{method}

\subsection{Problem Statement and Method Overview}

\begin{figure*}[h]
\centering
\includegraphics[scale=0.50]{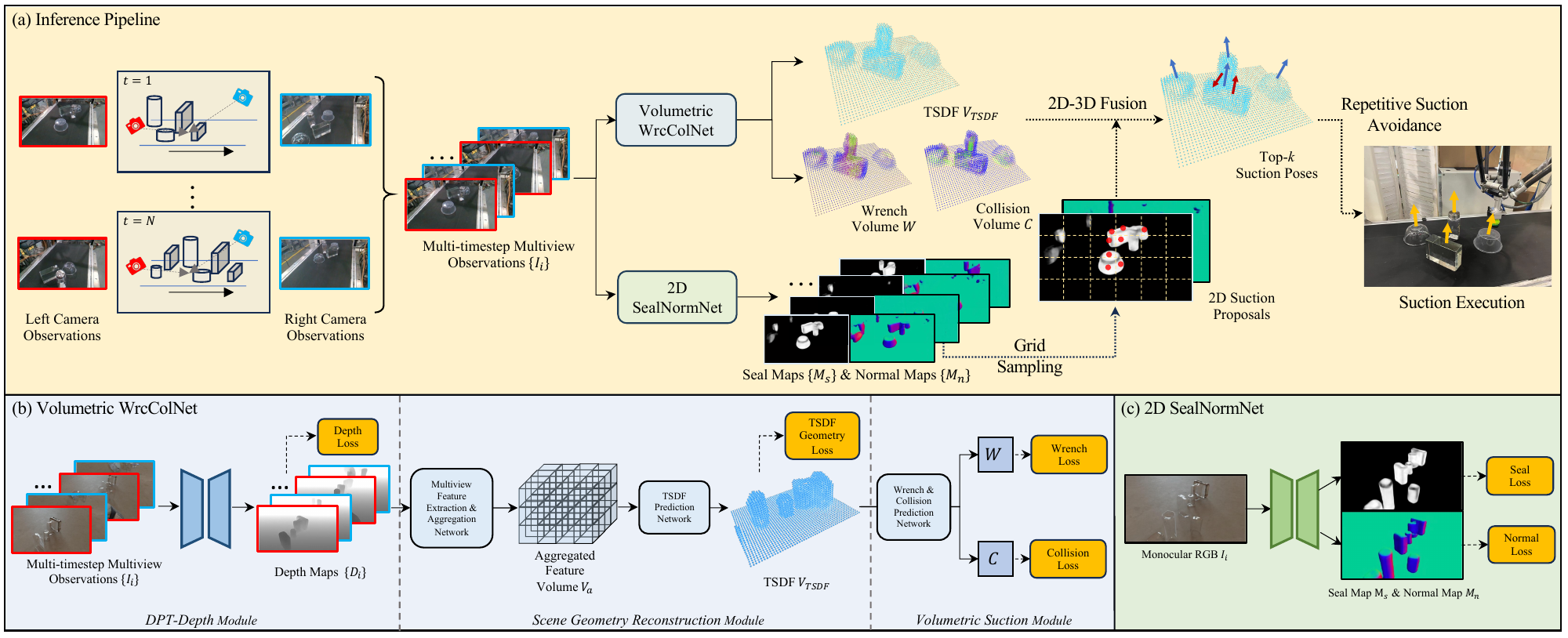}
\vspace{-8pt}
\caption{The framework of our proposed STOPNet: 
(a) \textbf{Inference Pipeline}: multi-timestep multiview RGB images are captured by two cameras as the input to our framework. Volumetric TSDF, wrench and collision volumes are predicted from our \textit{Volumetric WrcColNet}, while seal and normal maps are predicted from our \textit{2D SealNormNet}. We then use the grid sampling method to sample proposal suctions from predicted 2D seal maps and conduct 2D-3D fusion to obtain Top-\textit{k} suction poses for robot execution.
(b) \textbf{Volumetric WrcColNet}: from RGB input we predict depth maps, from which we extract and aggregate features to construct the aggregated feature volume $V_a$. Then we predict the TSDF volume $V_{TSDF}$ and conduct volumetric wrench-collision predictions;
(c) \textbf{2D SealNormNet}: from monocular RGB images, we jointly predict seal and normal maps.
}
\vspace{-18pt}
\label{pipeline_framework}
\end{figure*}

Given a sequence of RGB images of objects on a production line, the goal of the proposed robotic system is to detect the 6-DoF suction poses and then execute the suction to remove objects on the line, as shown in Fig.~\ref{fig/teaser}. The production line moves at a known fixed speed from left to right. On the left side is the reconstruction zone, where RGB images are captured by two cameras at regular timesteps. Then 3D reconstruction and suction pose detection are conducted in real time and the final suction pose at this timestep will be sent to the robot arm. When objects move into the robot's workspace on the right side, \textit{i.e.}, the suction zone, the robot arm will execute the suction based on a position shift. The process is pipelined and repeated until there is no detected valid suction pose.

After the production line is stable, for each timestep, we get the paired images of this timestep, and retrieve paired images of previous $N-1$ timesteps, obtaining $N$ pairs of multi-timestep images in total. In this way, we formulate the 6-DoF suction detection as a learning problem that maps a set of paired RGB images from $N$ timesteps, \textit{i.e.}\ $2N$ images, ${\{I_i\}}_{i=1,\dots,2N}$, to a set of 6-DoF suction poses ${\{S_j|S_j=(p_j,d_j,s_j)\}}$, where, for each detected suction $S_j$, $p_j \in \mathbb{R}^3$ is the suction position, $d_j \in \mathbb{R}^3$ is the suction direction, and $s_j \in [0,1]$ is the overall suction score, which is composed of 3 parts: seal score, wrench score and collision score, where seal score $s_{seal} \in [0,1]$ models the seal formation of the suction cup, wrench score $s_{wrench} \in [0,1]$ models the resistency to the wrench caused by gravity, and the binary collision score $s_{collision} \in \{0,1\}$ denotes whether a suction pose is in collision with other objects in the scene. The overall score for a suction pose is then computed by:
\vspace{-1pt}
\begin{equation}
    \label{eq/score}
    s_j = s_{seal} \times s_{wrench} \times s_{collision}
\end{equation}

Given that the production line is moving at a fixed and known speed, we can transform the observed multi-timestep multiview images into static multiview images. This is because we can consider the objects to be stationary and the cameras to be moving, as illustrated in Fig.~\ref{fig/teaser}. It allows us to calculate the transformed camera extrinsics of each timestep based on the production line's speed and original camera calibration. We can then use these images, along with the transformed camera extrinsics, to perform 3D reconstruction and suction pose detection.

Our proposed framework is composed of two branches, \textit{Volumetric WrcColNet} and \textit{2D SealNormNet}, as illustrated in Fig.~\ref{pipeline_framework}. For each timestep, in \textit{Volumetric WrcColNet}, we leverage our \textit{DPT-Depth Module} to conduct monocular depth estimation on separate RGB inputs and feed depth predictions into a multiview \textit{Scene Geometry Reconstruction Module} to output volumetric TSDF of the reconstruction zone. Taking TSDF as input, the following \textit{Volumetric Suction Module} learns to predict grid-level wrench scores and collision scores of the whole scene. In \textit{2D SealNormNet}, we take RGB images as input to jointly predict pixel-level seal scores and surface normals. The two branches are separately trained and fused together at inference time to output the Top-\textit{k} suction poses for each timestep.

\subsection{Volumetric WrcColNet: Scene Reconstruction and Wrench-Collision Prediction}

To better detect suction poses on a moving production line, it is crucial to wholly and accurately reconstruct the scene geometry in a generalizable way. Therefore our proposed \textit{Volumetric WrcColNet}, which is composed of \textit{DPT-Depth Module}, \textit{Scene Geometry Reconstruction Module} and \textit{Volumetric Suction Module}, utilizes multi-timestep multiview images to reconstruct the whole 3D scene to learn better suction detection. All the modules above are jointly trained in a fully end-to-end manner.

\subsubsection{DPT-Depth Module}

Since real-world data are expensive and labor intensive, it is more common to get training data from simulation environments, thus the sim2real gap is an inevitable issue to tackle. Under these circumstances, we propose to first learn a module to unify synthetic data and real data into the depth domain.

Dense Prediction Transformers, \textit{i.e.}\ DPT~\cite{ranftl2021vision} is a large-scale vision transformer that can be applied to many dense image-to-image prediction tasks, such as semantic segmentation, monocular depth estimation, \textit{etc}. DPT's network architecture contains an encoder that is composed of a series of embedding layers, transformer blocks, reassemble layers and fusion layers to encode the RGB input into dense image feature maps, and a convolutional decoder network to output dense prediction results of the same resolution as input.

Specifically, we leverage a DPT model pretrained on \textit{MIX 6}~\cite{ranftl2021vision}, a large-scale real-world monocular depth estimation dataset containing about 1.4 million images, and finetune it in our setting. Therefore the network inherits the abundant real-world knowledge in advance and thus better generalizes to the real world. Taking multi-timestep RGB observations $\{I_i\}$ as input, the module outputs depth map predictions $\{D_i\}$ for 3D reconstruction. In this way we discard redundant textures and illuminations from input and unify input domains into the depth domain, which helps to bridge the sim2real gap. To supervise the \textit{DPT-Depth Module}, an L1 pixel-level \textbf{Depth Loss $L_d$} is utilized to directly supervise depth estimations.

\subsubsection{Scene Geometry Reconstruction Module}

Inspired by~\cite{Dai2023GraspNeRF}, first we propose a \textit{Multiview Feature Extraction and Aggregation Network} to encode the depth predictions into a volumetric representation. To be specific, we divide our 3D reconstruction zone into grids of size $X$, $Y$ and $Z$ and utilize a CNN encoder to extract input depth prediction feature maps. Each grid point $x \in \mathbb{R}^{XYZ}$ is then projected to the $2N$ depth feature maps and queries the corresponding features, which are then aggregated among views using MLPs to form an aggregated feature volume $V_a \in \mathbb{R}^{XYZ \times C}$.

After obtaining aggregated feature volume $V_a$, similar to~\cite{wang2023neus} and~\cite{Dai2023GraspNeRF}, we propose a \textit{TSDF Prediction Network} to predict volumetric TSDF $V_{TSDF} \in [-1,1]^{XYZ}$ from $V_a$, supervised by a L1 \textbf{TSDF Geometry Loss $L_g$}.



\subsubsection{Volumetric Suction Module}

Motivated by~\cite{breyer2020volumetric}, taking predicted TSDF as input, we propose a 3D CNN encoder-decoder \textit{Wrench and Collision Prediction Network} to predict volumetric wrench scores $W \in [0,1]^{XYZ}$ and collision scores $C \in \{0,1\}^{XYZ}$. We explicitly supervise the
L2 \textbf{Wrench Loss} $L_w$ and binary cross entropy \textbf{Collision Loss} $L_c$ on grids.


Therefore the whole \textit{Volumetric WrcColNet}'s objective $L_{3D}$ can be formulated as:

\vspace{-1pt}
\begin{equation}
    L_{3D} = \alpha_1 L_{d} + \alpha_2 L_{g} + \alpha_3 L_{w} + \alpha_4 L_{c} 
\end{equation}

\subsection{2D SealNormNet: Suction Pose Detection}

As SuctionNet-1Billion~\cite{suctionnet} indicates, each suction pose contains 6 degrees of freedom and can be evaluated through three dimensions: seal, wrench and collision. Wrench and collision mainly focus on the global characteristics of objects and scenes, while seal heavily relies on the local characteristics of object surfaces. Therefore a prediction of seal scores from unreliable 3D representations can do great harm to the performance, and so do surface normals, which are the default choices of suction directions. To this end, we propose to predict seal scores and surface normal in a 2D manner.

Specifically, taking RGB as input, we utilize a DPT model that shares the transformer encoder layers to extract image features and leverage different convolutional decoder layers to jointly predict the seal map $M_s$ and the normal map $M_n$. And we directly supervise the dense pixel loss, which is:

\vspace{-1pt}
\begin{equation}
    L_{2D} = \beta_1\| M_s - M^*_s \|_2 + \beta_2 \| M_n - M^*_n \|_1
\end{equation}

where $M^*_s$ and $M^*_n$ are the GT seal and normal map.

\subsection{Inference Pipeline}

At inference time, we need to conduct 2D-3D fusion to acquire Top-\textit{k} suction poses to execute. First we do grid sampling on 2D seal maps to get pixel positions of high seal scores and query corresponding normals on predicted normal maps to obtain our 2D suction proposals. Based on the predicted TSDF volume $V_{TSDF}$, we can render the corresponding de-noised depth maps using the Marching Cube Algorithm~\cite{conf/siggraph/LorensenC87}. This allows us to obtain 3D positions $\{p_j\}$ and normals $\{d_j\}$ from 2D pixels. Given 3D points of seal scores $s_{seal} \in [0,1]$, both wrench scores $s_{wrench} \in [0,1]$ and collision scores $s_{collision} \in \{0,1\}$ of the nearest grids from the predicted wrench and collision volumes will be assigned to these points. Finally, we calculate the overall scores by Eq.~\ref{eq/score} and rank them to get $\{s_j\}$. In this way we obtain the 3D Top-\textit{k} suction poses ${\{S_j|S_j=(p_j,d_j,s_j)\}}$, which can be further executed by robot suction cups.

As Top-\textit{k} suction poses are unaware of object identities, multiple suctions can be predicted on the same object. To avoid this, we propose \textit{Repetitive Suction Avoidance}, where we predict instance masks with MobileSAM~\cite{mobile_sam} and apply heuristic filters to ensure each object is only sucked once. At each timestep, if there exist valid and non-repetitive suction poses, one final suction pose will be selected and sent to the robot arm. It will be executed once the object appears in the end-effector’s workspace.

\subsection{Synthetic Data Generation and Sim2real Gap Reduction}

Since our case is specific, no existing dataset can satisfy our needs. And as real-world data with annotations can be expensive and labor intensive, training on synthetic datasets is more costless and scalable. Hence, motivated by~\cite{Dai2023GraspNeRF}, we propose an automatic data generation pipeline and a large-scale production line suction dataset, containing over 1000 production line layouts, 10000 multiview scenes, 300K images and 40M suction poses of diverse objects.

To generate the data, we first collect a wide range of CAD models and textures. Then we randomly place these objects on a moving production line in PyBullet simulator~\cite{coumans2021}. Based on simulation output, we render photo-realistic multiview RGB images, depth and normal maps in Blender~\cite{blender}. For suction annotation, we use APIs provided by~\cite{suctionnet} to generate GT seal, wrench and collision scores. We then render the corresponding 2D seal maps using PyTorch3D~\cite{ravi2020pytorch3d}. Example data can be found in Fig.~\ref{fig/teaser} upper right.

To better bridge the sim2real gap, we leverage domain randomization to randomize object arrangements, illuminations, camera poses and backgrounds. Combining this with our finetuned \textit{DPT-Depth Module}, we largely reduce sim2real gaps, as models trained on our synthetic dataset can easily generalize to the real world and achieve higher performance than other methods, which will be covered in Section \ref{exp}.

\section{EXPERIMENTS} \label{exp}

\begin{table*}[h]
\setlength{\abovecaptionskip}{-5pt}
\setlength{\belowcaptionskip}{-0pt}
\caption{Evaluation for different methods in the simulation environment \\ Each cell: transparent / mixed }
\label{simulation_exp}
\begin{center}
\begin{tabular}{|c|c|c|c|c|c|c|c|}
\hline
\multicolumn{2}{|c|}{Metrics} & TSDF MAE $\downarrow$ & Surf. TSDF MAE $\downarrow$ & Seal MAE $\downarrow$ & Collision Acc. (\%) $\uparrow$ & AP@Top-1 $\uparrow$ & AP@Top-5 $\uparrow$ \\
\hline
\multirow{3}*{Seen} & DPT-Fusion~\cite{ranftl2021vision} & 4.73 / 2.88 & 20.03 / 12.25 & \textbf{0.018} / \textbf{0.020} & 93.93 / 94.44 & 0.55 / 0.44 & 0.55 / 0.42 \\
~ & SuctionNeRF~\cite{Dai2023GraspNeRF} & 0.88 / 0.71 & 3.09 / 2.03 & 0.280 / 0.269 & 95.89 / 95.25 & 0.05 / 0.04 & 0.05 / 0.04 \\
\cline{2-8}
~ & Ours & \textbf{0.55} / \textbf{0.47} & \textbf{2.08} / \textbf{1.43} & \textbf{0.018} / \textbf{0.020} & \textbf{96.32} / \textbf{95.68} & \textbf{0.65} / \textbf{0.80} & \textbf{0.58} / \textbf{0.71} \\
\hline

\multirow{3}*{Similar} & DPT-Fusion~\cite{ranftl2021vision} & 6.12 / 3.78 & 23.32 / 12.43 & \textbf{0.025} / \textbf{0.024} & 93.36 / 94.60 & 0.45 / 0.55 & 0.30 / 0.56 \\
~ & SuctionNeRF~\cite{Dai2023GraspNeRF} & 1.41 / 1.01 & 4.02 / 2.52 & 0.237 / 0.222 & 95.29 / 95.69 & 0.00 / 0.10 & 0.10 / 0.07 \\
\cline{2-8}
~ & Ours & \textbf{0.97} / \textbf{0.65} & \textbf{2.82} / \textbf{1.59} & \textbf{0.025} / \textbf{0.024} & \textbf{95.57} / \textbf{96.51} & \textbf{0.55} / \textbf{0.85} & \textbf{0.59} / \textbf{0.81} \\
\hline

\multirow{3}*{Novel} & DPT-Fusion~\cite{ranftl2021vision} & 3.30 / 3.23 & 18.77 / 16.62 & \textbf{0.032} / \textbf{0.034} & 93.08 / 93.06 & 0.42 / 0.51 & 0.22 / 0.43 \\
~ & SuctionNeRF~\cite{Dai2023GraspNeRF} & 0.67 / 0.85 & 2.43 / 3.14 & 0.394 / 0.371 & 95.56 / 93.20 & 0.00 / 0.10 & 0.06 / 0.07 \\
\cline{2-8}
~ & Ours & \textbf{0.47} / \textbf{0.59} & \textbf{1.84} / \textbf{2.19} & \textbf{0.032} / \textbf{0.034} & \textbf{95.87} / \textbf{93.34} & \textbf{0.55} / \textbf{0.80} & \textbf{0.44} / \textbf{0.65} \\
\hline
\end{tabular}
\end{center}
\vspace{-18pt}
\end{table*}

In this section, we evaluate the performance of our proposed method for production line suction tasks in both simulation and real-world environments. We also perform ablation studies in the real world to analyze the effectiveness of different modules of our framework.

\subsection{Implementation Details}

As indicated in Section \ref{method}, we set the size of the reconstruction zone grid as $X=50,Y=40,Z=30$ in $cm$, and the number of timesteps used in one reconstruction is $N=5$. We train our model for a maximum number of 300K iterations utilizing the Adam optimizer~\cite{kingma2017adam} with a learning rate of $1\times10^{-4}$ and a learning rate decay of 0.5 per 100K iterations.

\subsection{Experiment Setup}

\subsubsection{Object Sets}

In simulation, we utilize 160 hand-scale objects from~\cite{jiang2021synergies} and assign them with transparent textures. We then replenish the object set with 80 package objects of similar scales and assign them with diverse real-world textures in order to learn suction detection for practical industrial usage. Following~\cite{graspnet} and~\cite{suctionnet}, objects are split into seen, similar, and novel sets. We train our model only on the seen set and test them on all three sets in novel arrangements to check the generalizability of our framework.

In the real world, we gather around 30 different types of diverse objects, including transparent ones and other common items, to put on the production line. All the objects are unseen to our model so we split them into similar and novel sets based on their geometry, as shown in Fig.~\ref{fig/teaser}.

\subsubsection{Real Robot Setup}

As in Fig.~\ref{fig/teaser}, we set two RealSense D415 cameras diagonally to capture RGB images (depth is not used in our method) and set the speed of the production line to 100mm/s. For motion planning, since the objects are moving at a fixed speed, we utilize a heuristic algorithm that moves the robot arm above to track the suction position in the workspace. When the two trajectories coincide, the suction cup will push down to execute the suction.

For the final suction pose selection from Top-\textit{k} proposals at each timestep, we take heights and directions of suction poses into account to adapt 6-DoF suction poses to our 4-DoF AtomRobot D3P-1100-P3 parallel robot arm. We set the timestep length to 1s, and the inference time of our whole pipeline is around 0.5s using a single GPU, leading to continuous and pipelined suctions on the production line in 1Hz, which is compatible with our robot arm's cycle time.

\subsection{Baseline Methods}

We compare our method with the following baselines in simulation or real-world environments.

\begin{itemize}
    \item \textbf{DPT-Fusion}~\cite{ranftl2021vision}. As multiview depths can be directly used to fuse a TSDF in a non-learning way, we utilize the toolbox provided by~\cite{zeng20163dmatch} to fuse a TSDF volume from predicted depths by finetuned DPT~\cite{ranftl2021vision} in place of our \textit{Scene Geometry Reconstruction Module}. We utilize our \textit{2D SealNormNet} for seal and normal prediction.
    
    \item \textbf{SuctionNeRF}~\cite{Dai2023GraspNeRF}. Based on GraspNeRF~\cite{Dai2023GraspNeRF}, a multiview 3D reconstruction and grasp detection framework using generalizable NeRF, we modify its VGN~\cite{breyer2020volumetric} grasping detection head layer to predict seal, wrench and collision scores to adapt to our setting.
    
    \item \textbf{SuctionNet}~\cite{suctionnet}. An object suction detection framework that takes RGBD images as input. It gets a partial point cloud from depth input and a 2D heatmap from RGBD input, and then executes suction by the heatmap scores and projected point coordinates.

    \item \textbf{2D Object Detector}~\cite{Jocher_YOLO_by_Ultralytics_2023}. In industry, a common practice is to use a 2D object detector to detect objects' centroids. To ensure generalizability, we utilize YOLOv8~\cite{Jocher_YOLO_by_Ultralytics_2023} for object detection and finetuned DPT~\cite{ranftl2021vision} for depth estimation.
\end{itemize}

\subsection{Simulation Experiments}

\subsubsection{Metrics}

In simulation environments, we mainly focus on different methods' performance on 3D object reconstruction and 6-DoF suction prediction precision, so we compare our method with DPT-Fusion~\cite{ranftl2021vision} and SuctionNeRF~\cite{Dai2023GraspNeRF} using multiple metrics, including the \textbf{TSDF MAE} and \textbf{Surface TSDF MAE} (converted to millimeters) to evaluate reconstruction quality of the whole scene and object surfaces, \textbf{Seal MAE} to evaluate seal score prediction and \textbf{Collision Accuracy} to evaluate collision score prediction. Following~\cite{suctionnet}, we also evaluate the overall suction score prediction using \textbf{AP@Top-\textit{k}}, the average of $\rm{AP}_s$ with threshold $\rm{s}$ ranging from 0.2 to 0.8, with an interval of $\delta_{\rm{s}} = 0.2$. We report the AP value under Top-1 and Top-5.

\subsubsection{Results and Analysis} \label{sim_analysis}

As presented in Table \ref{simulation_exp}, in the simulation environment, our method outperforms all the baselines for all combinations of object sets (seen/similar/novel) and object types (transparent/mixed), demonstrating the superiority of our method.

Compared to non-learning DPT-Fusion, we perform significantly better due to our method's great improvement in reconstruction quality, which verifies that our \textit{Scene Geometry Reconstruction Module} can significantly reduce the inconsistency of multiple monocular depth estimations.

Also, compared to SuctionNeRF, our method improves reconstruction quality because of the prior knowledge on depth estimation embodied in the pretrained \textit{DPT-Depth Module}. Moreover, better reconstruction quality benefits suction collision detection, as our method has better performance on collision accuracy.

\begin{table}[h]
\setlength{\abovecaptionskip}{-15pt}
\caption{Evaluation for different methods in the real world \\ Each cell: transparent / mixed }
\label{real_world_exp}
\begin{center}
\resizebox{\linewidth}{!}{
\begin{tabular}{|c|c|c|c|}
\hline
\multicolumn{2}{|c|}{Metrics} & SR (\%) $\uparrow$ & DR (\%) $\uparrow$ \\
\hline
\multirow{5}*{Similar} & DPT-Fusion~\cite{ranftl2021vision} & 13.76 / 27.64 & 12.50 / 26.25\\
~ & SuctionNeRF~\cite{Dai2023GraspNeRF} & 48.78 / 42.56 & 46.87 / 39.33\\
~ & SuctionNet~\cite{suctionnet} & 27.27 / 40.00 & 22.72 / 33.33\\
~ & 2D Object Detector~\cite{Jocher_YOLO_by_Ultralytics_2023} & 27.77 / 28.57 & 14.28 / 15.38\\
\cline{2-4}
~ & Ours & \textbf{87.64} / \textbf{90.38} & \textbf{70.90} / \textbf{78.77}\\
\hline
\multirow{5}*{Novel} & DPT-Fusion~\cite{ranftl2021vision} & 6.81 / 14.01 & 6.38 / 12.71\\
~ & SuctionNeRF~\cite{Dai2023GraspNeRF} & 19.56 / 13.15 & 19.14 / 11.36\\
~ & SuctionNet~\cite{suctionnet} & 9.66 / 15.38 & 8.10 / 12.50\\
~ & 2D Object Detector~\cite{Jocher_YOLO_by_Ultralytics_2023} & 10.00 / 11.76 & 10.00 / 9.75\\
\cline{2-4}
~ & Ours & \textbf{69.23} / \textbf{85.52} & \textbf{61.36} / \textbf{73.86}\\
\hline
\end{tabular}
}
\end{center}
\vspace{-12pt}
\end{table}

For seal prediction and AP@Top-\textit{k}, our method greatly surpasses SuctionNeRF because instead of taking flawed TSDF as input, we predict seal scores in a 2D manner, which takes RGB as input so that the model can better learn local characteristics of object surfaces, which is crucial for seal scores. Our method also beats DPT-Fusion since our reconstruction quality is better.

In addition, our method's improvement on similar and novel object sets demonstrates that our method has better generalizability due to our pretrained \textit{DPT-Depth Module} and using predicted depth as input of 3D reconstruction.

\subsection{Real Robot Experiments}

\begin{figure}
\centering
\includegraphics[width=1.0\linewidth]{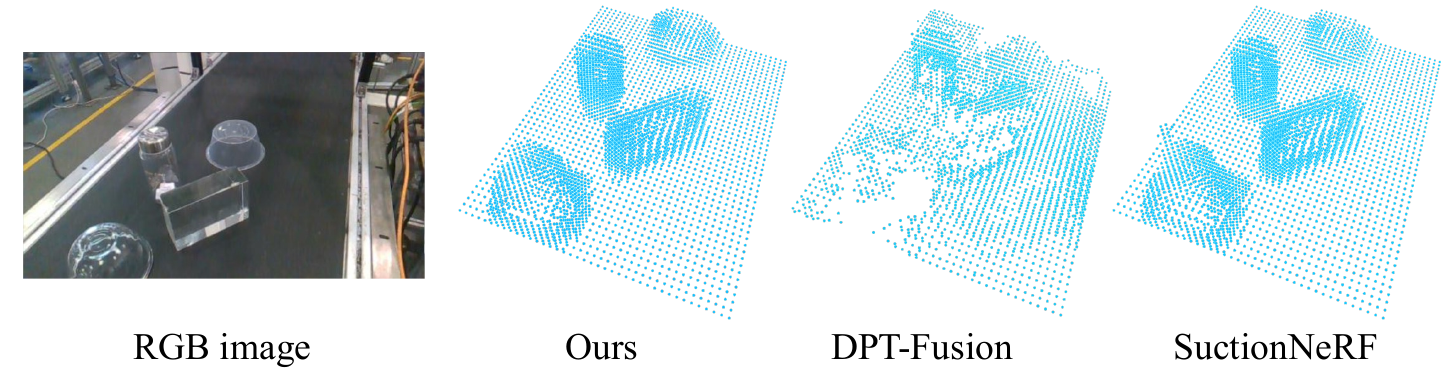}
\vspace{-15pt}
\caption{Visualization of TSDF reconstructed by different methods. The geometry from DPT-Fusion, which is directly fused from depth estimations has severe flaws, indicating that our method can greatly reduce the inconsistency of multiple monocular depth estimations. In comparison to SuctionNeRF, which uses direct RGB input and produces floaters in 3D reconstruction, our method results in more accurate reconstruction and better suction detection.
}
\vspace{-15pt}
\label{fig/visualization}
\end{figure}

We evaluate our method and all baselines in the real world with a production line and objects configured as in Fig.~\ref{fig/teaser}. Each experiment consists of 3 rounds with each round featuring 120 objects which are sequentially placed in the same arrangement. A suction is considered successful if it removes the object from the production line.

\subsubsection{Metrics}

As the GT labels are not available in the real world, following\cite{Dai2023GraspNeRF}, we measure the performance by \textbf{Success Rate (SR)}: the ratio of the successful suction number to the attempt number; and \textbf{Declutter Rate (DR)}: the average percentage of removed objects to all objects. 

\subsubsection{Results and Analysis}
\label{real_analysis}

Table \ref{real_world_exp} presents SR \& DR of real-world object suction on the production line with the same objects and arrangements. Results show that our method beats all the baselines for all combinations of object sets (similar/novel) and object types (transparent/mixed). The great improvement in the novel object set also indicates our method's generalizability. In addition, in Fig.~\ref{fig/visualization} we provide real-world reconstruction results using different methods.

Based on the results, it is clear that our approach performs much better than DPT-Fusion and SuctionNeRF, as explained in Section~\ref{sim_analysis}. Additionally, the visualization in Fig.~\ref{fig/visualization} demonstrates that our method can create more precise and superior 3D scenes in real-world settings. This is in contrast to other methods that generate artifacts or floaters that hinder wrench prediction and collision detection.

Compared to SuctionNet, which needs depth for input, our method performs significantly better in SR and DR, especially in transparent objects, indicating that errors in transparent object depths greatly harm the performance of RGBD-based methods.

Finally, our method outperforms the basic 2D object detector method because we leverage multi-timestep images to wholly reconstruct the scene, making it more capable of handling cluttered scenes. Also, rather than just selecting the centroids of objects, our method can better predict seal scores on objects, contributing to better performance.

However, we also notice some failure cases. For example, occasionally there will be a small offset in reconstruction and suction detection because of the unstable condition of production lines.

\subsection{Ablation Studies}

To analyze our method design, we conduct ablation studies on different configurations in the real world. As the effectiveness of our \textit{2D SealNormNet} has been analyzed at Section~\ref{sim_analysis}, this section focuses mainly on our \textit{Volumetric WrcColNet}. Results can be found in Table~\ref{ablation_real}.

\begin{table}[h]
\setlength{\abovecaptionskip}{-15pt}
\caption{Ablation studies in the real world \\ Each cell: transparent / mixed }
\label{ablation_real}
\begin{center}
\resizebox{\linewidth}{!}{
\begin{tabular}{|c|c|c|c|}
\hline
\multicolumn{2}{|c|}{Metrics} & SR (\%) $\uparrow$ & DR (\%) $\uparrow$ \\
\hline
\multirow{4}*{Similar} & RGB & 80.37 / 74.71 & 65.64 / 65.19\\
~ & w/o pretraining & 68.33 / 79.60 & 58.15 / 72.62\\
~ & 1/5 Data & 70.12 / 76.43 & 49.09 / 62.82\\
\cline{2-4}
~ & Ours & \textbf{87.64} / \textbf{90.38} & \textbf{70.90} / \textbf{78.77}\\
\hline
\multirow{4}*{Novel}  & RGB & 50.94 / 52.08 & 48.21 / 42.01\\
~ & w/o pretraining & 50.00 / 51.85 & 43.63 / 41.17\\
~ & 1/5 Data & 60.00 / 70.00 & 50.00 / 57.14\\
\cline{2-4}
~ & Ours & \textbf{69.23} / \textbf{85.52} & \textbf{61.36} / \textbf{73.86}\\
\hline
\end{tabular}
}
\end{center}
\vspace{-15pt}
\end{table}

First, we examine the effect of taking depth maps as input to our \textit{Scene Geometry Reconstruction Module}. We change the module's input from predicted depth maps to raw RGB images, and the performance degradation indicates that removing redundant elements, such as textures and illuminations from input helps to bridge the sim2real gap.

Then we analyze the effect of the pretrained \textit{DPT-Depth Module} on \textit{MIX 6}~\cite{ranftl2021vision} dataset. Results show that compared with training \textit{DPT-Depth Module} from scratch, our method can better generalize to the real world due to the prior knowledge from pretraining.

To show the necessity of the large scale of our synthetic dataset, we cut down training data to 1/5 of the original data. The resulting decrease in performance highlights the significance of a large training data scale for model performance.



\section{CONCLUSIONS}

In this work, we propose STOPNet, a multiview RGB-based 3D reconstruction and 6-DoF suction detection framework for transparent objects on production lines. We also present a large-scale synthetic dataset and corresponding data generation pipeline for object suction on production lines. We also leverage multiple practices to reduce the sim2real gap. Experiments in simulation and the real world demonstrate our method's superiority over current methods and great generalizability to the real world. We hope our work can benefit related research and industrial practices.










\newpage

\bibliographystyle{IEEEtran}

\bibliography{IEEEabrv,mybibfile}

\begin{thebibliography}{10}
\providecommand{\url}[1]{#1}
\csname url@rmstyle\endcsname
\providecommand{\newblock}{\relax}
\providecommand{\bibinfo}[2]{#2}
\providecommand\BIBentrySTDinterwordspacing{\spaceskip=0pt\relax}
\providecommand\BIBentryALTinterwordstretchfactor{4}
\providecommand\BIBentryALTinterwordspacing{\spaceskip=\fontdimen2\font plus
\BIBentryALTinterwordstretchfactor\fontdimen3\font minus \fontdimen4\font\relax}
\providecommand\BIBforeignlanguage[2]{{%
\expandafter\ifx\csname l@#1\endcsname\relax
\typeout{** WARNING: IEEEtran.bst: No hyphenation pattern has been}%
\typeout{** loaded for the language `#1'. Using the pattern for}%
\typeout{** the default language instead.}%
\else
\language=\csname l@#1\endcsname
\fi
#2}}

\bibitem{cad_matching}
L.~Duc~Hanh, N.~Luat, and L.~Bich, ``Combining 3d matching and image moment based visual servoing for bin picking application,'' \emph{International Journal on Interactive Design and Manufacturing (IJIDeM)}, vol.~16, 03 2022.

\bibitem{industry4.0}
\BIBentryALTinterwordspacing
S.~D’Avella, C.~A. Avizzano, and P.~Tripicchio, ``Ros-industrial based robotic cell for industry 4.0: Eye-in-hand stereo camera and visual servoing for flexible, fast, and accurate picking and hooking in the production line,'' \emph{Robotics and Computer-Integrated Manufacturing}, vol.~80, p. 102453, 2023. [Online]. Available: \url{https://www.sciencedirect.com/science/article/pii/S0736584522001351}
\BIBentrySTDinterwordspacing

\bibitem{assembly}
M.~Pfeffer, C.~Goth, D.~Craiovan, and J.~Franke, ``3d-assembly of molded interconnect devices with standard smd pick \& place machines using an active multi axis workpiece carrier,'' in \emph{2011 IEEE International Symposium on Assembly and Manufacturing (ISAM)}, 2011, pp. 1--6.

\bibitem{pick_by_vision}
\BIBentryALTinterwordspacing
R.~Reif and W.~A. G{\"{u}}nthner, ``Pick-by-vision: augmented reality supported order picking,'' \emph{Vis. Comput.}, vol.~25, no. 5-7, pp. 461--467, 2009. [Online]. Available: \url{https://doi.org/10.1007/s00371-009-0348-y}
\BIBentrySTDinterwordspacing

\bibitem{morph}
M.~Naeem, S.~Aslam, M.~Suhaib, S.~Gul, Z.~Murtaza, and M.~J. Khan, ``Design and implementation of pick and place manipulation system for industrial automation,'' in \emph{2021 International Conference on Artificial Intelligence and Mechatronics Systems (AIMS)}, 2021, pp. 1--6.

\bibitem{practical}
\BIBentryALTinterwordspacing
K.~Castelli, A.~M.~A. Zaki, and H.~Giberti, ``Development of a practical tool for designing multi-robot systems in pick-and-place applications,'' \emph{Robotics}, vol.~8, no.~3, 2019. [Online]. Available: \url{https://www.mdpi.com/2218-6581/8/3/71}
\BIBentrySTDinterwordspacing

\bibitem{akinola2021dynamic}
I.~Akinola, J.~Xu, S.~Song, and P.~K. Allen, ``Dynamic grasping with reachability and motion awareness,'' in \emph{2021 IEEE/RSJ International Conference on Intelligent Robots and Systems (IROS)}, 2021, pp. 9422--9429.

\bibitem{andy-affordance}
\BIBentryALTinterwordspacing
A.~Zeng, S.~Song, K.-T. Yu, E.~Donlon, F.~R. Hogan, M.~Bauza, D.~Ma, O.~Taylor, M.~Liu, E.~Romo, N.~Fazeli, F.~Alet, N.~C. Dafle, R.~Holladay, I.~Morona, P.~Q. Nair, D.~Green, I.~Taylor, W.~Liu, T.~Funkhouser, and A.~Rodriguez, ``Robotic pick-and-place of novel objects in clutter with multi-affordance grasping and cross-domain image matching,'' \emph{The International Journal of Robotics Research}, vol.~41, no.~7, pp. 690--705, 2022. [Online]. Available: \url{https://doi.org/10.1177/0278364919868017}
\BIBentrySTDinterwordspacing

\bibitem{mahler2018dexnet}
J.~Mahler, M.~Matl, X.~Liu, A.~Li, D.~Gealy, and K.~Goldberg, ``Dex-net 3.0: Computing robust vacuum suction grasp targets in point clouds using a new analytic model and deep learning,'' in \emph{2018 IEEE International Conference on Robotics and Automation (ICRA)}, 2018, pp. 5620--5627.

\bibitem{dexnet4.0}
\BIBentryALTinterwordspacing
J.~Mahler, M.~Matl, V.~Satish, M.~Danielczuk, B.~DeRose, S.~McKinley, and K.~Goldberg, ``Learning ambidextrous robot grasping policies,'' \emph{Science Robotics}, vol.~4, no.~26, p. eaau4984, 2019. [Online]. Available: \url{https://www.science.org/doi/abs/10.1126/scirobotics.aau4984}
\BIBentrySTDinterwordspacing

\bibitem{suctionnet}
H.~Cao, H.-S. Fang, W.~Liu, and C.~Lu, ``Suctionnet-1billion: A large-scale benchmark for suction grasping,'' \emph{IEEE Robotics and Automation Letters}, vol.~6, no.~4, pp. 8718--8725, 2021.

\bibitem{RGBD-seg-sucdetect}
Y.~Fu, T.~Sun, L.~Wang, S.~Yu, L.~Deng, B.~Chen, L.~Li, Y.~Xie, S.~Deng, and H.~Yin, ``Rgb-d instance segmentation-based suction point detection for grasping,'' in \emph{2022 IEEE International Conference on Robotics and Biomimetics (ROBIO)}, 2022, pp. 1643--1650.

\bibitem{sajjan2019cleargrasp}
S.~Sajjan, M.~Moore, M.~Pan, G.~Nagaraja, J.~Lee, A.~Zeng, and S.~Song, ``Clear grasp: 3d shape estimation of transparent objects for manipulation,'' in \emph{2020 IEEE International Conference on Robotics and Automation (ICRA)}, 2020, pp. 3634--3642.

\bibitem{fang2022transcg}
H.~Fang, H.-S. Fang, S.~Xu, and C.~Lu, ``Transcg: A large-scale real-world dataset for transparent object depth completion and a grasping baseline,'' \emph{IEEE Robotics and Automation Letters}, vol.~7, no.~3, pp. 7383--7390, 2022.

\bibitem{dai2022domain}
Q.~Dai, J.~Zhang, Q.~Li, T.~Wu, H.~Dong, Z.~Liu, P.~Tan, and H.~Wang, ``Domain randomization-enhanced depth simulation and restoration for perceiving and grasping specular and transparent objects,'' in \emph{Computer Vision -- ECCV 2022}, S.~Avidan, G.~Brostow, M.~Ciss{\'e}, G.~M. Farinella, and T.~Hassner, Eds.\hskip 1em plus 0.5em minus 0.4em\relax Cham: Springer Nature Switzerland, 2022, pp. 374--391.

\bibitem{Weng-2020-123091}
T.~Weng, A.~Pallankize, Y.~Tang, O.~Kroemer, and D.~Held, ``Multi-modal transfer learning for grasping transparent and specular objects,'' \emph{IEEE Robotics and Automation Letters}, vol.~5, no.~3, pp. 3791--3798, 2020.

\bibitem{soft}
H.~Cao, J.~Huang, Y.~Li, J.~Zhou, and Y.~Liu, ``Fuzzy-depth objects grasping based on fsg algorithm and a soft robotic hand,'' in \emph{2021 IEEE/RSJ International Conference on Intelligent Robots and Systems (IROS)}, 2021, pp. 3948--3954.

\bibitem{zhou2019glassloc}
Z.~Zhou, T.~Pan, S.~Wu, H.~Chang, and O.~C. Jenkins, ``Glassloc: Plenoptic grasp pose detection in transparent clutter,'' in \emph{2019 IEEE/RSJ International Conference on Intelligent Robots and Systems (IROS)}, 2019, pp. 4776--4783.

\bibitem{ghostpose}
J.~Chang, M.~Kim, S.~Kang, H.~Han, S.~Hong, K.~Jang, and S.~Kang, ``Ghostpose: Multi-view pose estimation of transparent objects for robot hand grasping,'' in \emph{2021 IEEE/RSJ International Conference on Intelligent Robots and Systems (IROS)}, 2021, pp. 5749--5755.

\bibitem{Dai2023GraspNeRF}
Q.~Dai, Y.~Zhu, Y.~Geng, C.~Ruan, J.~Zhang, and H.~Wang, ``Graspnerf: Multiview-based 6-dof grasp detection for transparent and specular objects using generalizable nerf,'' in \emph{2023 IEEE International Conference on Robotics and Automation (ICRA)}, 2023, pp. 1757--1763.

\bibitem{suction_cup_modeling}
J.~Hudoklin, S.~Seo, M.~Kang, H.~Seong, A.~T. Luong, and H.~Moon, ``Vacuum suction cup modeling for evaluation of sealing and real-time simulation,'' \emph{IEEE Robotics and Automation Letters}, vol.~7, no.~2, pp. 3616--3623, 2022.

\bibitem{Lasi2014}
\BIBentryALTinterwordspacing
H.~Lasi, P.~Fettke, H.-G. Kemper, T.~Feld, and M.~Hoffmann, ``Industry 4.0,'' \emph{Business {\&} Information Systems Engineering}, vol.~6, no.~4, pp. 239--242, Aug 2014. [Online]. Available: \url{https://doi.org/10.1007/s12599-014-0334-4}
\BIBentrySTDinterwordspacing

\bibitem{ranftl2021vision}
R.~Ranftl, A.~Bochkovskiy, and V.~Koltun, ``Vision transformers for dense prediction,'' in \emph{Proceedings of the IEEE/CVF International Conference on Computer Vision (ICCV)}, October 2021, pp. 12\,179--12\,188.

\bibitem{wang2023neus}
\BIBentryALTinterwordspacing
P.~Wang, L.~Liu, Y.~Liu, C.~Theobalt, T.~Komura, and W.~Wang, ``Neus: Learning neural implicit surfaces by volume rendering for multi-view reconstruction,'' in \emph{Advances in Neural Information Processing Systems}, M.~Ranzato, A.~Beygelzimer, Y.~Dauphin, P.~Liang, and J.~W. Vaughan, Eds., vol.~34.\hskip 1em plus 0.5em minus 0.4em\relax Curran Associates, Inc., 2021, pp. 27\,171--27\,183. [Online]. Available: \url{https://proceedings.neurips.cc/paper_files/paper/2021/file/e41e164f7485ec4a28741a2d0ea41c74-Paper.pdf}
\BIBentrySTDinterwordspacing

\bibitem{breyer2020volumetric}
\BIBentryALTinterwordspacing
M.~Breyer, J.~J. Chung, L.~Ott, R.~Siegwart, and J.~Nieto, ``Volumetric grasping network: Real-time 6 dof grasp detection in clutter,'' in \emph{Proceedings of the 2020 Conference on Robot Learning}, ser. Proceedings of Machine Learning Research, J.~Kober, F.~Ramos, and C.~Tomlin, Eds., vol. 155.\hskip 1em plus 0.5em minus 0.4em\relax PMLR, 16--18 Nov 2021, pp. 1602--1611. [Online]. Available: \url{https://proceedings.mlr.press/v155/breyer21a.html}
\BIBentrySTDinterwordspacing

\bibitem{conf/siggraph/LorensenC87}
\BIBentryALTinterwordspacing
W.~E. Lorensen and H.~E. Cline, ``Marching cubes: A high resolution 3d surface construction algorithm.'' in \emph{SIGGRAPH}, M.~C. Stone, Ed.\hskip 1em plus 0.5em minus 0.4em\relax ACM, 1987, pp. 163--169. [Online]. Available: \url{http://dblp.uni-trier.de/db/conf/siggraph/siggraph1987.html#LorensenC87}
\BIBentrySTDinterwordspacing

\bibitem{mobile_sam}
C.~Zhang, D.~Han, Y.~Qiao, J.~U. Kim, S.-H. Bae, S.~Lee, and C.~S. Hong, ``Faster segment anything: Towards lightweight sam for mobile applications,'' \emph{arXiv preprint arXiv:2306.14289}, 2023.

\bibitem{coumans2021}
E.~Coumans and Y.~Bai, ``Pybullet, a python module for physics simulation for games, robotics and machine learning,'' \url{http://pybullet.org}, 2016--2021.

\bibitem{blender}
\BIBentryALTinterwordspacing
B.~O. Community, \emph{Blender - a 3D modelling and rendering package}, Blender Foundation, Stichting Blender Foundation, Amsterdam, 2018. [Online]. Available: \url{http://www.blender.org}
\BIBentrySTDinterwordspacing

\bibitem{ravi2020pytorch3d}
N.~Ravi, J.~Reizenstein, D.~Novotny, T.~Gordon, W.-Y. Lo, J.~Johnson, and G.~Gkioxari, ``Accelerating 3d deep learning with pytorch3d,'' \emph{arXiv:2007.08501}, 2020.

\bibitem{kingma2017adam}
\BIBentryALTinterwordspacing
D.~P. Kingma and J.~Ba, ``Adam: {A} method for stochastic optimization,'' in \emph{3rd International Conference on Learning Representations, {ICLR} 2015, San Diego, CA, USA, May 7-9, 2015, Conference Track Proceedings}, Y.~Bengio and Y.~LeCun, Eds., 2015. [Online]. Available: \url{http://arxiv.org/abs/1412.6980}
\BIBentrySTDinterwordspacing

\bibitem{jiang2021synergies}
\BIBentryALTinterwordspacing
Z.~Jiang, Y.~Zhu, M.~Svetlik, K.~Fang, and Y.~Zhu, ``Synergies between affordance and geometry: 6-dof grasp detection via implicit representations,'' in \emph{Robotics: Science and Systems XVII, Virtual Event, July 12-16, 2021}, D.~A. Shell, M.~Toussaint, and M.~A. Hsieh, Eds., 2021. [Online]. Available: \url{https://doi.org/10.15607/RSS.2021.XVII.024}
\BIBentrySTDinterwordspacing

\bibitem{graspnet}
H.-S. Fang, C.~Wang, M.~Gou, and C.~Lu, ``Graspnet-1billion: A large-scale benchmark for general object grasping,'' in \emph{Proceedings of the IEEE/CVF Conference on Computer Vision and Pattern Recognition (CVPR)}, June 2020.

\bibitem{zeng20163dmatch}
A.~Zeng, S.~Song, M.~Niessner, M.~Fisher, J.~Xiao, and T.~Funkhouser, ``3dmatch: Learning local geometric descriptors from rgb-d reconstructions,'' in \emph{Proceedings of the IEEE Conference on Computer Vision and Pattern Recognition (CVPR)}, July 2017.

\bibitem{Jocher_YOLO_by_Ultralytics_2023}
\BIBentryALTinterwordspacing
G.~Jocher, A.~Chaurasia, and J.~Qiu, ``{YOLO by Ultralytics},'' Jan. 2023. [Online]. Available: \url{https://github.com/ultralytics/ultralytics}
\BIBentrySTDinterwordspacing

\end{thebibliography}

\end{document}